\documentclass{article}
\usepackage{ijcai25}

\usepackage{times}
\usepackage{soul}
\usepackage{url}
\usepackage[hidelinks]{hyperref}
\usepackage[utf8]{inputenc}
\usepackage[small]{caption}
\usepackage{graphicx}
\usepackage{amsmath}
\usepackage{amsthm}
\usepackage{booktabs}
\usepackage{algorithm}
\usepackage{algorithmic}
\usepackage{multirow} \usepackage{makecell}
\usepackage[switch]{lineno}
\usepackage[utf8]{inputenc} 
\usepackage[T1]{fontenc}    
\usepackage{url}            
\usepackage{booktabs}       
\usepackage{amsfonts}       
\usepackage{nicefrac}       
\usepackage{microtype}      
\usepackage{xcolor}         
\usepackage{graphicx}
\usepackage{multirow} 
\usepackage{enumitem}
\usepackage{amsmath}
\usepackage{picinpar}
\usepackage{bm}

\usepackage{graphicx}
\usepackage{subcaption}

\title{OpenCarbon: A Contrastive Learning-based Cross-Modality Neural Approach for High-Resolution Carbon Emission Prediction Using Open Data}

\author{
Jinwei Zeng$^1$\and
Yu Liu$^1$ \and
Guozhen Zhang$^2$\and
Jingtao Ding$^1$ \and  
Yuming Lin$^1$ \and
Jian Yuan$^1$ \and
Yong Li\thanks{Corresponding author.}$^1$  \\
\affiliations
$^1$Tsinghua University\\
$^2$TsingRoc\\
\emails
zengjw17@gmail.com,
liyong07@tsinghua.edu.cn
}

\begin{document}

\maketitle

\begin{abstract}
Accurately estimating high-resolution carbon emissions is crucial for effective emission governance and mitigation planning. While conventional methods for precise carbon accounting are hindered by substantial data collection efforts, the rise of open data and advanced learning techniques offers a promising solution. Once an open data-based prediction model is developed and trained, it can easily infer emissions for new areas based on available open data. To address this, we incorporate two modalities of open data, satellite images and point-of-interest (POI) data, to predict high-resolution urban carbon emissions, with satellite images providing macroscopic and static and POI data offering fine-grained and relatively dynamic functionality information. However, estimating high-resolution carbon emissions presents two significant challenges: the intertwined and implicit effects of various functionalities on carbon emissions, and the complex spatial contiguity correlations that give rise to the agglomeration effect. Our model, OpenCarbon, features two major designs that target the challenges: a cross-modality information extraction and fusion module to extract complementary functionality information from two modules and model their interactions, and a neighborhood-informed aggregation module to capture the spatial contiguity correlations. Extensive experiments demonstrate our model's superiority, with a significant performance gain of 26.6\% on $R^2$. Further generalizability tests and case studies also show OpenCarbon's capacity to capture the intrinsic relation between urban functionalities and carbon emissions, validating its potential to empower efficient carbon governance and targeted carbon mitigation planning. Codes and data are available: \url{https://github.com/JinweiZzz/OpenCarbon}. 
\end{abstract}

\section{Introduction}
Accurately accounting for high-resolution urban carbon emissions has become an increasingly critical issue~\cite{stechemesser2012carbon}. Urban carbon emissions now account for more than 70\% of global emissions~\cite{agreement2015paris,worldbank} and continue to rise. To fully understand the sources of these emissions and effectively design carbon-reduction policies targeting high-emission areas, it is crucial to account for urban carbon emissions at a high resolution~\cite{stechemesser2012carbon}.

Despite the importance of emission accounting, traditional carbon accounting methods face a trade-off between accuracy and the effort required for data acquisition~\cite{cai2018china,olivier19991990,dai2016closing}. Existing approaches to carbon emission accounting generally fall into two categories: bottom-up and top-down~\cite{hutchins2017comparison,bohringer2008combining,bohringer1998synthesis}. The bottom-up approach, which relies on point emission statistics from sensors or fine-grained activity data~\cite{gurney2020vulcan,gurney2009high}, is accurate but requires extensive data collection, making it unsustainable. In contrast, the top-down approach distributes regional sectoral emission totals---calculated from fuel consumption statistics---across high-resolution areas using proxy factors~\cite{olivier1994emission}. However, the ideal assumption of positive correlations between proxies and emissions often leads to inaccuracies. Given the strengths of machine learning and deep learning in identifying complex correlations, along with the vast availability of open data---which is widely used in socioeconomic prediction tasks~\cite{yeh2020using,aiken2022machine,xu2020ar2net,wang2016crime}---developing a model that predicts carbon emissions using open data presents a promising solution. Once trained, the model can accurately infer a region’s emissions using readily available open data, maintaining accuracy while significantly reducing costly data collection efforts.

\begin{figure}[t]
    \centering
    \includegraphics[width=0.95\columnwidth]{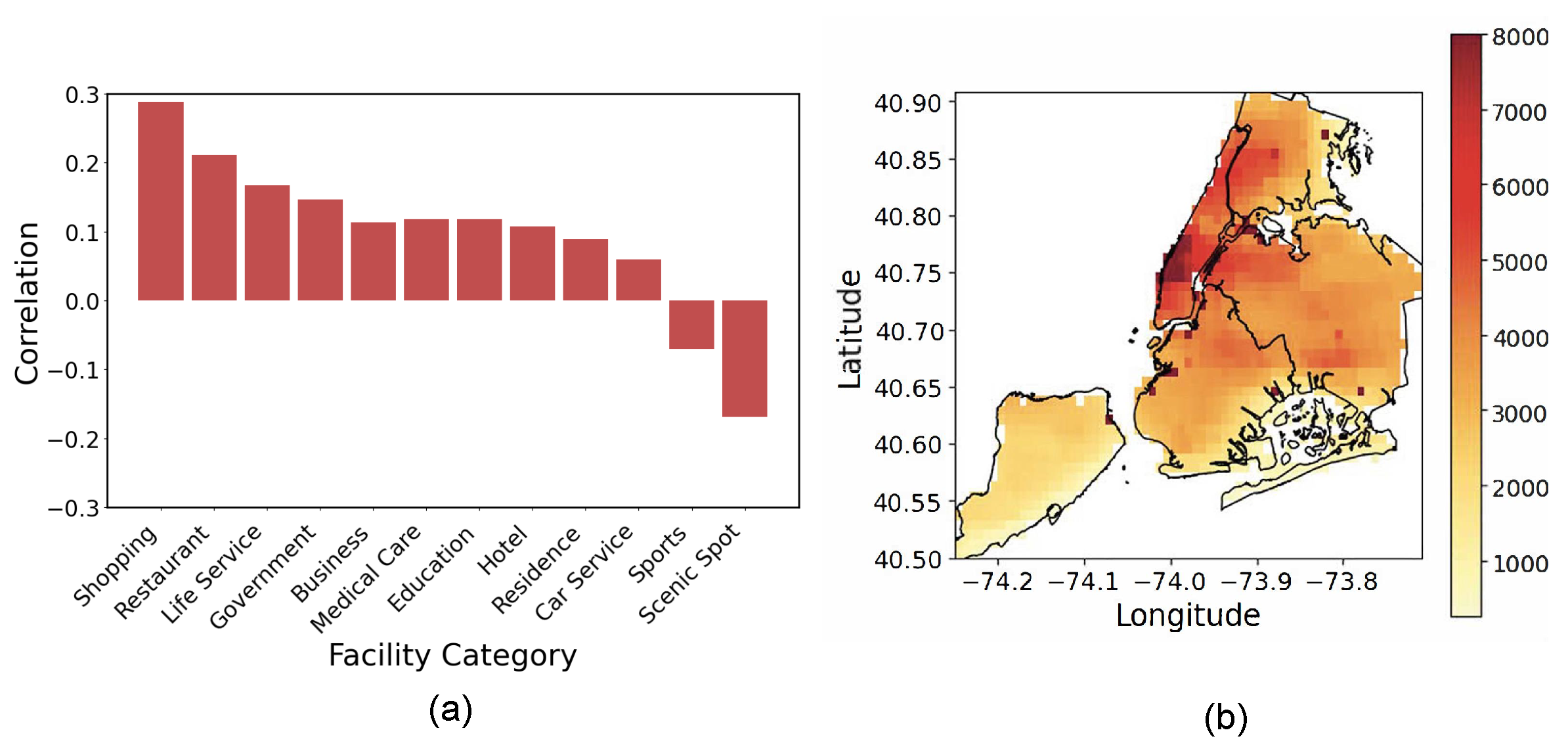}
    \caption{Illustrations of the (a)functional effect and (b)spatial agglomeration effect of urban carbon emission distribution. (a) Correlation between the shares of different POI types and carbon emissions in Beijing. (b) Spatial distribution of New York's carbon emissions. }
    \label{fig:intro}
\end{figure}

Nevertheless, directly applying existing open data-based socioeconomic prediction methods to predict carbon emissions fails to capture the unique and challenging characteristics of carbon emission spatial distribution: the functional effect and spatial agglomeration effect. While existing socioeconomic indicators typically reflect one type of activity or urban functionality, carbon emissions result from the combination of diverse functionalities that have distinct effects on emissions~\cite{dhakal2009urban,liu2020carbon} (Fig~\ref{fig:intro}(a)). For instance, traffic roads generate transportation carbon emissions as vehicles burn petrol~\cite{wang2015carbon}, while residential areas involve heating and cooking activities that consume natural gas and liquefied petroleum gas~\cite{nejat2015global,zhang2015household}. Learning the implicit and coupled relationships between such functionalities and carbon emissions is complex and challenging. Meanwhile, urban carbon emissions exhibit a strong and unique spatial agglomeration effect, originating from the continuity in city functionality layouts~\cite{han2018effects,wang2018agglomeration,wang2020effect}. As shown in Figure~\ref{fig:intro}(b), adjacent areas in New York City have similar emission levels, and the overall carbon emission distribution in the city exhibits continuity. Characterizing such continuity is challenging but essential.

To model the unique characteristics of carbon emissions, we incorporate two types of open data—satellite images and point-of-interest (POI) data—to predict high-resolution spatial carbon emissions for two main reasons. First, satellite images and POI data provide complementary insights into a region's functionalities. Satellite images offer a relatively macroscopic yet comprehensive overview of how the city's static functional zones are distributed~\cite{xu2020ar2net,zhao2022identifying}, while POI data captures the fine-grained dynamic intensity of human activities, reflecting the rapid and continuous updates of facilities within urban functional zones~\cite{fan2018understanding}. Second, while satellite images convey information about the spatial layout of each functional zone, POI data is typically geo-tagged and contains valuable location-specific information. Together, these data types form the foundation for modeling the spatial correlations between urban functionalities and activities. Therefore, our primary goal is to effectively leverage these two data types to model the unique and challenging aspects of estimating high-resolution urban carbon emissions.

To bridge the gap, we propose our model, \textbf{OpenCarbon}, to predict high-resolution urban \underline{carbon} emissions with \underline{open} data of satellite images and POI data. To model the coupled and implicit functional effects from the complementary open data modalities, we propose a cross-modality information extraction and fusion module. This module first extracts information from each modality, explicitly modeling the distinct effects of different functionalities using a function-dimension-wise attention mechanism. It then extracts complementary functional information across the two modalities through a contrastive loss design, aiming to enable each modality to borrow underrepresented information from the other. To capture the spatial contiguity of functionality layouts underlying the spatial agglomeration effect, OpenCarbon features a neighborhood-informed agglomeration modeling module that uses convolutional layers to extract the global context of the neighborhood and a cross-attention mechanism to model grid-neighborhood interactions. Our contribution may be summarized into three parts: 

\begin{itemize}
    \item We tackle the critical challenge of data collection in high-accuracy, high-resolution carbon accounting by developing an open data-based model that provides accurate carbon estimation while significantly reducing the data collection burden. 
    \item Recognizing the unique functional and spatial agglomeration effects of urban carbon emissions, we propose a robust neighborhood-informed attentive neural network with contrastive learning for cross-modality fusion that achieves complementary modality extraction and spatial contiguity modeling. 
    \item Extensive experiments on three large-scale real-world datasets validate our model's superiority over existing state-of-the-art methods, with an average increase of 26.6\% in terms of $R^2$, showing our model's potential to facilitate convenient high-resolution carbon estimation and the carbon governance that follows. 
\end{itemize}

\section{Related Works}
\subsection{Conventional High-resolution Carbon Emission Accounting}
High-resolution accounting of carbon emissions is essential for effective, targeted carbon emission governance. Traditional carbon emission accounting methods are generally classified into two categories: bottom-up and top-down approaches~\cite{bohringer2008combining}. Bottom-up methods aggregate monitored point emission data to the desired resolution~\cite{gurney2020vulcan,gurney2009high}, but they require extensive data collection and are not easily scalable across large areas. In contrast, top-down methods distribute regional carbon emissions, typically calculated from regional fuel consumption statistics, to the target resolution using proxy factors~\cite{olivier1994emission}, such as population. However, because carbon emissions are influenced by a range of factors, relying solely on proxy factors in top-down approaches often leads to inaccuracies. Therefore, we conclude that conventional high-resolution carbon accounting methods face a dilemma between accuracy and the substantial effort required for data collection.

\subsection{Carbon Emission Prediction with Open Data}
With the development of machine learning and deep learning and the boom in open data, researchers made a large trial in incorporating them to relieve the data collection pressure~\cite{yang2020modeling,lu2017predicting,chen2024spatiotemporal,wu2022high}. Yang et al.~\cite{yang2020modeling} ensembled several multi-layer perception models to predict city-level emissions with nighttime light satellite imagery. Chen et al.~\cite{chen2024spatiotemporal} constructed a graph neural network to model the spatiotemporal dependencies between nearby counties or districts, predicting temporally fine-grained emissions using available monthly emissions statistics. However, most of these methods target region-level or city-level carbon emissions and may not be suitable for high-resolution prediction due to the limited resolution of the data they involve, or because they inadequately characterize fine-grained functional and spatial interaction information. Therefore, we conclude that the potential of high-resolution open data for predicting carbon emissions remains underdeveloped.


\subsection{Socioeconomic Prediction with Satellite Images and POI Distribution Data}
Since our solution uses satellite images and POI distribution data as inputs, we review existing works that leverage these data types. Most studies incorporating POI data represent POIs as a distribution count vector. Wang et al.\cite{wang2016crime} developed a neural network model to predict neighborhood crime rates using the POI vector, while Yu et al.\cite{yu2018smartphone} used POI counts to predict smartphone application usage. For satellite image-related works, visual encoder networks have been applied to learn task-specific image representations~\cite{yeh2020using,perez2017poverty}. Researchers have also used unsupervised and self-supervised methods to create unified representations for various socioeconomic tasks. Jean et al.\cite{jean2019tile2vec} designed a triplet loss function to enhance representation similarity for nearby grids, and Xi et al.\cite{xi2022beyond} proposed similarity metrics for geo-adjacent grids and grids with similar POI distributions to improve grid semantics. However, to the best of our knowledge, no work has specifically explored the modeling potential of these two complementary data sources for high-resolution carbon emissions. The challenge of effectively extracting complementary information and modeling the implicit functional and spatial agglomeration effects remains underexplored.

\section{Preliminaries \& Problem Statement}
\subsection{Definitions}
As our work incorporates two types of open data, the satellite image, and POI distribution data, we provide their definition and collection sources here. \\
\noindent
\textbf{Satellite Image:}
Satellite images are overhead-view pictures of the Earth's surface taken by satellites orbiting the planet. The recent development of remote sensing has enabled the acquisition of ~\textbf{global, high-resolution, and safe} satellite images~\cite{campbell2011introduction}. Our data source is ArcGIS\footnote{https://www.arcgis.com/home}, whose maximum pixel resolution is 1.19m. 

\noindent
\textbf{POI Distribution:}
Point-of-interest(POI) data is the geo-tagged facility distribution data that is widely available in various map services and data providers. Some crowd-sourcing map service providers, such as OpenStreetMap, provide access to the global POI data~\cite{zhou2022assessing}. In this study, we use public POI data with location and category information provided from SafeGraph\footnote{https://docs.safegraph.com/docs/places} and Tencent Map.

Notably, all two data modalities are sourced from publicly available, privacy-protected sources with no security risks.

\subsection{Problem Statement}
We target urban carbon emissions at a spatial granularity that has been paid much attention to in carbon emission governance, the 1km $\times$ 1km grids. For each grid $i$ in the grid set $\mathcal{A}$, given its corresponding satellite image and POI distribution data, our task is to predict its yearly carbon emission. 


\section{Method: OpenCarbon}
\begin{figure*}[ht]
    \centering
    \includegraphics[width=1.7\columnwidth]{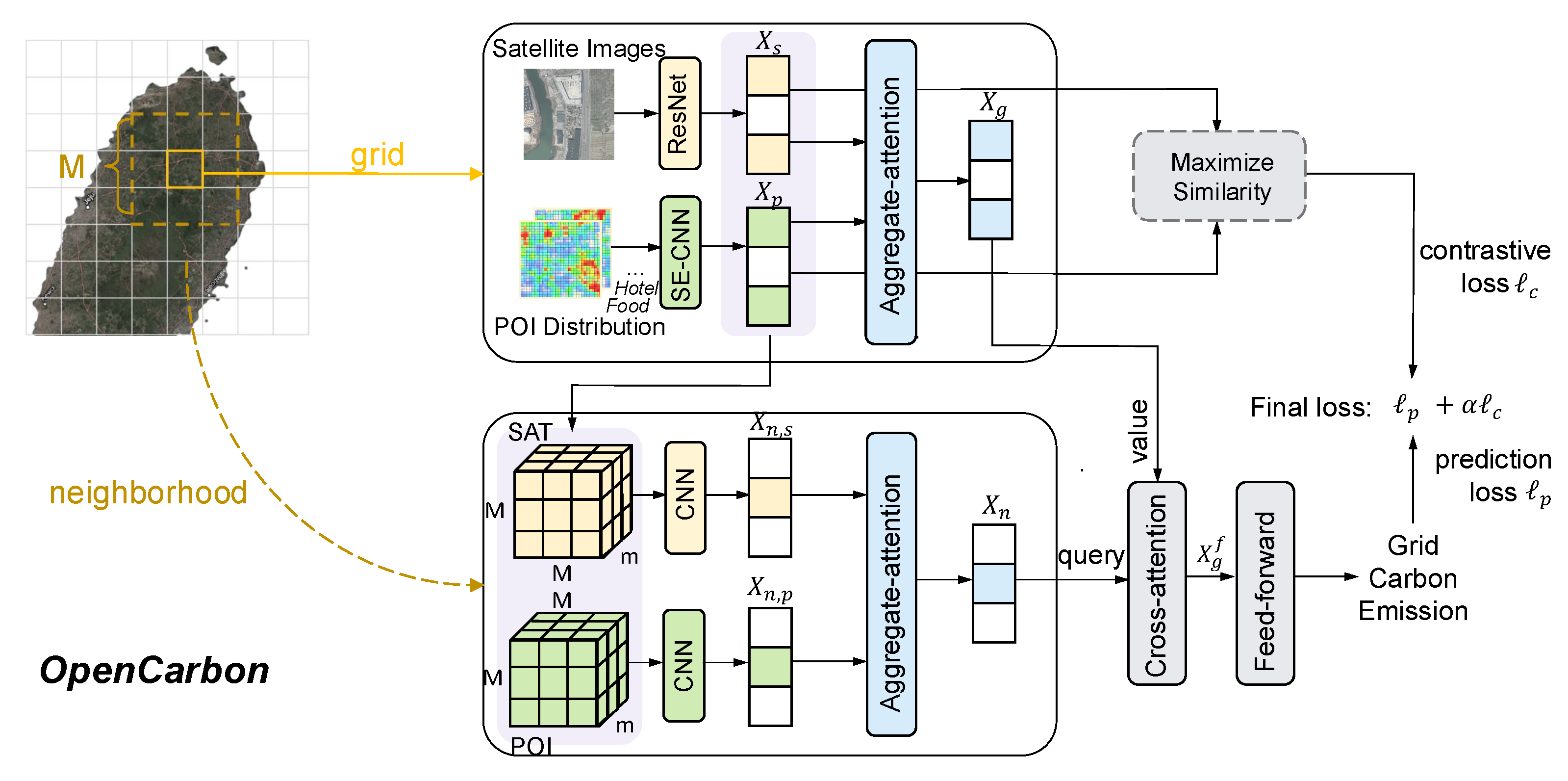}~\label{fig:all}
    \caption{The overall framework of our OpenCarbon model.}
    \label{fig:backbone}
\end{figure*}
In this section, we present our method, OpenCarbon (Figure~\ref{fig:backbone}), which predicts high-resolution carbon emissions using open data. OpenCarbon comprises two main components: (1) a cross-modality information extraction and fusion module that leverages complementary functional information from two modalities and models their interactions, and (2) a neighborhood-informed aggregation module designed to capture the continuity of city functionality layouts, thereby modeling the agglomeration effect. In the following sections, we provide a detailed introduction to these two modules and the training process.

\subsection{Cross-modality Information Extraction and Fusion}
\subsubsection{Single-modality Representation Learning}
Satellite images offer a broad overview of the functional layouts and their scale, whereas POI data provides complementary, detailed insights into the functionality and activity intensity within each layout. For example, while satellite images provide an immediate sense of a shopping mall's scale, they offer little insight into the composition of its internal store categories. POI data supplements this understanding. Therefore, aligning the observation scales of both modalities is crucial, allowing our model to effectively match corresponding information and develop a comprehensive representation of the area's functionalities.

We use satellite image pixels as the fundamental unit of granularity to organize the geo-tagged POI data into a three-dimensional matrix with consistent pixel dimensions. The first two dimensions represent the spatial distribution of the POIs, while the third encodes their types. The matrix is constructed by mapping each POI to its corresponding matrix element based on its position within the grid. In this way, we obtain the grid-level matrix inputs of the satellite images and POI distribution.

For the satellite image modality, given its visual nature, we incorporate the ResNet-18~\cite{he2016deep} architecture to obtain the embedding. Since the POI matrix is relatively sparse, we apply multiple layers of CNNs to obtain a representation of the POI distribution. However, since carbon emissions result from the aggregation of various activities, with distinct and implicit relationships between each activity type and its corresponding emission, we introduce a squeeze-and-excitation (SE) block~\cite{hu2018squeeze} into every CNN layer to account for functionality heterogeneity in the POI distribution representation.

Specifically, for any input $X \in \mathbb{R}^{H \times W \times C}$ to the CNN layer, a global average pooling layer serves as the squeeze function to capture the global information of each channel, which can be formulated as

\begin{equation} z_{c} = \frac{1}{H \times W} (\sum_{h}\sum_w X_{hwc}). \end{equation}

An excitation function is then introduced to generate attention weights for different channels, scaling the input along the function dimension. The process can be expressed as follows:

\begin{equation} s_c = {\rm Sigmoid}(W_2\cdot {\rm ReLU}(W_1 z_c)), \end{equation}

\begin{equation} X'{\cdot, \cdot, c} = s_c \cdot X{\cdot, \cdot, c}, \quad c=1, 2, ..., C. \end{equation}

Here, $W_1 \in \mathbb{R}^{\frac{C}{n} \times C}$ and $W_2 \in \mathbb{R}^{C \times \frac{C}{n}}$ are the learnable projection matrices, and $n$ represents the tunable compression ratio.

\subsubsection{Cross-modality Complementary Information Fusion}
Once each modality has been represented individually, the next step is to enable each modality to learn complementary information from the other and effectively fuse the two. To accomplish this, we employ contrastive learning, aiming to allow each modality to borrow and encode missing or underrepresented information from the other. Specifically, denoting the input from the satellite image modality as $X_s$ and that from the POI modality as $X_p$, we introduce the well-acknowledged contrastive learning loss, NT-Xent (Normalized Temperature-scaled Cross Entropy) loss~\cite{chen2020simple}, which can be formulated as 

\begin{align*}
\ell_{c, i} = & -\log \frac{\exp\left(\frac{\operatorname{sim}(\mathbf{X}_{s, i}, \mathbf{X}_{p, i})}{\tau}\right)}{\sum_{k=1}^{N} {1}_{[k \neq i]} \exp\left(\frac{\operatorname{sim}(\mathbf{X}_{s, i}, \mathbf{X}_{p, k})}{\tau}\right)} \\
& -\log \frac{\exp\left(\frac{\operatorname{sim}(\mathbf{X}_{s, i}, \mathbf{X}_{p, i})}{\tau}\right)}{\sum_{k=1}^{N} {1}_{[k \neq i]} \exp\left(\frac{\operatorname{sim}(\mathbf{X}_{s, k}, \mathbf{X}_{p, i})}{\tau}\right)}.
\label{equ:contrastive}
\end{align*}
Here \(\operatorname{sim}(\mathbf{x}_i, \mathbf{x}_j)\) represents the cosine similarity between \(\mathbf{x}_i\) and \(\mathbf{x}_j\). \(\tau\) is the temperature parameter, used to adjust the distribution of similarity values. \(N\) represents the number of samples in a batch.

After obtaining the complementary representations, we apply the aggregate-attention mechanism to adjust the weights of the two modalities during the representation fusion. Specifically, the mechanism maps the representations to a score space and adaptively learns the scores so that the more informative modality receives a higher weight. The attention weights are computed as:

\begin{equation}
\begin{aligned}
   & m_{k} = a\cdot {\rm Tanh}(W\cdot X_k+ b), k \in \{s, p\}, \\
   & s_{k} = \frac{{\rm exp}({m_k})}{\sum_{k \in \{s, i\}} {\rm exp}({m_k})}, k \in \{s, i\}.
\end{aligned}
\end{equation}
Here $W$ is a learnable matrix, and $a$ and $b$ are learnable vectors. The fused grid-level representation $X_{g}$ is 
\begin{equation}
  X_{g} = \sum_{k \in \{s, i\}} {s_k \cdot X_k}
  \label{equ:fusion}
\end{equation}

\subsection{Neighborhood-informed Agglomeration Modeling}
While a graph structure can represent grid adjacency, it does not capture the relative position information between each grid, as the neighbors of a graph node are unordered. Therefore, we turn to convolutional layers to model the global representation of the neighborhood. Specifically, in single-modality representation learning, we obtain grid-level representations for satellite images \(X_s \in \mathbb{R}^{m}\) and POI data \(X_p \in \mathbb{R}^{m}\). Considering the \(M \times M\) grids surrounding the target grid as the contextual range (where \(M\) is a hyperparameter), we concatenate the representations of the two modalities into two matrices, \(X_{n, s}\) and \(X_{n,p}\), each of size \(\mathbb{R}^{M \times M \times m}\). Each matrix is passed through several convolutional layers to learn the spatial agglomeration correlations of its neighborhood. The global representations are then aggregated using the previously introduced aggregate-attention mechanism. This process results in a neighborhood-level representation $X_n$ of the grid.

To capture the spatial continuity underlying the spatial agglomeration effect, we introduce the standard cross-attention mechanism. We use the contextual neighborhood representation as the query vector to model the spatial correlations and interactions. In this way, we obtain the final grid representation \( X^{f}_g \), which can be formulated as:

\begin{equation}
    {X^{f}_g} = {\rm Attention}(query=X_n, value=X_g) + X_g.
    \label{equ:attention}
\end{equation}


\subsection{Prediction \& Training}
By inputting the final grid representations $X^{f}_g$ into feed-forward layers, we get the predicted grid carbon emission $\hat{Y}$. During training, we employ the well-acknowledged mean absolute error loss as the prediction loss function, which can be formulated as 

\begin{equation}
   \ell_{p, i}  = ||\hat{y_i} -y_i||. 
   \label{equ: loss}
\end{equation}

The overall loss function is a weighted addition of the contrastive learning loss and the prediction loss:

\begin{equation}
   \mathcal{L}  = \sum_{i\in \mathcal{A}}{\ell_{p, i} + \alpha \cdot \ell_{c, i}}. 
   \label{equ: overall loss}
\end{equation}
where $\mathcal{A}$ is the grid set of our target region and $\alpha$ is a tunable coefficient. Notably, in implementation, we add the contrastive learning loss during training after 100 epochs to prevent information collapse between the two modalities in the early stages of training.
\section{Experimental Results}
\subsection{Experimental Setups}
\subsubsection{Datasets}
We select three representative regions with varying levels of development for a comprehensive evaluation: Greater London (UK), Beijing (China), and Yinchuan (China). The first two regions are relatively well-developed, yet they exhibit distinct industrial structures and emission levels. In contrast, Yinchuan is a relatively underdeveloped region. The sources and basic statistics of the three datasets are provided in Table~\ref{tbl:dataset}. The grid-level carbon emission statistics are all collected from ODIAC~\cite{oda2018open}, a well-acknowledged carbon emission inventory constructed using the hybrid means of point emission calculations and top-down allocations. Consistent with ODIAC, we set our prediction resolution as $1km$ $\times$ $1km$. 

\begin{table}[h]
    \centering
    \scalebox{0.9}{
    \begin{tabular}{l c c c}
        \toprule
        Region & Great London & Beijing & Yinchuan \\
        \midrule
        Area & $778\,km^2$ & $1381\,km^2$ & $475\,km^2$ \\
        GDP pc (\$) & $71k$ & $27k$ & $12k$ \\ 
        POI Source & SafeGraph & Map Service & Map Service \\
        Target Year & 2018 & 2018 & 2019 \\
        \bottomrule
    \end{tabular}
    }
    \caption{Summary statistics of our three main datasets.}~\label{tbl:dataset}
    \vspace{-5mm}
\end{table}

\begin{table*}[ht!]
    \centering
    \begin{tabular}{c c c c c c c c c c c }
        \toprule & &\multicolumn{3}{c}{London} &  
        \multicolumn{3}{c}{Beijing} &  \multicolumn{3}{c}{Yinchuan} \\
        \cmidrule{3-5} \cmidrule{6-8} \cmidrule{9-11}
        Groups & Models & $R^2$ & MAE & RMSE  & $R^2$ & MAE & RMSE  & $R^2$ & MAE & RMSE  \\
        \midrule
        \multirow{4}*{\makecell[c]{Carbon \\Prediction}} 
        & SVR & 0.395 & 0.312 & 0.389 & 0.460 & 0.396 & 0.513 & -0.167 & 0.664 & 0.803 \\
        & Stacked RFR & 0.310 & 0.328 & 0.415 & 0.476 & 0.399 & 0.505 & 0.014 & 0.611 & 0.738 \\
        & BPNN & -13.315 & 0.456 & 1.890 & 0.080 & 0.466 & 0.669 & -0.668 & 0.821 & 0.960 \\
        & CarbonGCN & -0.152 & 0.448 & 0.536 & -0.082 & 0.579 & 0.726 & -1.632 & 1.070 & 1.206 \\
        \cmidrule{1-11}
        \multirow{3}*{\makecell[c]{Satellite \\ Image \\ Representation}} 
        & READ & 0.338 & 0.328 & 0.406 & 0.558 & 0.364 & 0.464 & -0.095 & 0.633 & 0.778 \\
        & Tile2Vec & 0.527 & 0.269 & 0.343 & 0.567 & 0.352 & 0.459 & 0.070 & 0.598 & 0.717 \\
        & PG-SimCLR & \underline{0.612} & \underline{0.237} & \underline{0.311} & \underline{0.621} & \underline{0.334} & \underline{0.429} & \underline{0.444} & \underline{0.440} & \underline{0.554} \\
        \midrule
        \multirow{2}*{\makecell[c]{}} 
        & \textbf{OpenCarbon} & \bm{$0.786$} & \bm{$0.168$} & \bm{$0.231$} & \bm{$0.691$} & \bm{$0.302$} & \bm{$0.388$} & \bm{$0.622$} & \bm{$0.357$} & \bm{$0.457$} \\
        & \textbf{Improv.}   & \bm{$28.4\%$}  & \bm{$29.1\%$}  & \bm{$25.8\%$}  & \bm{$11.3\%$}  & \bm{$9.6\%$}   & \bm{$9.7\%$}   & \bm{$40.1\%$}  & \bm{$18.6\%$}  & \bm{$17.4\%$} \\
        \bottomrule
    \end{tabular}
    \caption{Performance comparisons on three main datasets. Training and testing datasets are split by regional divisions. The best results are in bold and the second-best results are underlined.}~\label{tbl:results}
\end{table*}

\subsubsection{Baselines}
We compare with both carbon emission prediction methods and satellite image-based socioeconomic indicator prediction methods. The carbon emission prediction methods typically incorporate related statistics to predict regional carbon emissions and we adapt them to the high-resolution grid level:

\begin{itemize}[leftmargin=*]
    \item \textbf{SVM}~\cite{mladenovic2016management}. Support vector machine method for emission prediction.
    \item \textbf{Stacked-RFR}~\cite{zhang2022estimating}. An ensemble two-layer stacked random forest regression model.
    \item \textbf{BPNN}~\cite{zhang2021towards}. A classical back propagation neural network. 
    \item \textbf{CarbonGCN}~\cite{chen2024spatiotemporal}. A graph neural network for neighborhood interaction modeling.
\end{itemize}
Since the inputs of these methods are vector-based, we transform satellite images into vectors by an acknowledged pretrained-encoder~\cite{han2020lightweight} and concatenate them with the POI count vector as input. Satellite image-based socioeconomic indicator prediction methods include:

\begin{itemize}[leftmargin=*] 
\item \textbf{READ}\cite{han2020lightweight}: A pretrained satellite image model using transfer learning on a large-scale partially-labeled dataset to learn robust and lightweight representations. 
\item \textbf{Tile2Vec}\cite{jean2019tile2vec}: Models the first law of geography by using triplet loss to maximize the similarity of geo-adjacent satellite image representations. 
\item \textbf{PG-SimCLR}~\cite{xi2022beyond}: Encourages geo-adjacent grids with similar facility distributions to have similar representations. 
\end{itemize} 
We adapt the unsupervised Tile2Vec and self-supervised PG-SimCLR methods into a supervised approach by adding a carbon emission prediction loss term and balancing the multiple losses. We also incorporate POI count vectors at the prediction stage for fair comparison.

\subsubsection{Metrics and Implementation}
To measure the prediction performance, we adopt three commonly used evaluation metrics: mean absolute error (MAE), rooted mean squared error (RMSE), and coefficient of determination ($R^2$)~\cite{xi2022beyond}. We performed a grid search on our hyperparameters, including learning rate, batch size, and neighborhood scale, and balancing coefficient. The search range for the neighborhood scale is \{3, 5, 7\}, and that for the balancing coefficient is \{1e-1, 1e-2, 1e-3\}. The hyperparameters of the baselines are all carefully grid searched. During training, we partition the dataset by \textbf{regional divisions}, selecting certain sub-regions for training and validation while reserving the remaining sub-regions for testing. As an example, for the Beijing dataset, we perform training and validation on five districts (including Xicheng and Chaoyang) and conduct testing on the Fengtai district. We perform experiment on an NVIDIA RTX 4090, with 12-20 GB GPU memory usage and < 4 hour training time. 

\subsection{Overall Performance}
We conduct extensive experiments on three distinct datasets to comprehensively evaluate the performance of OpenCarbon. From Table~\ref{tbl:results}, we can draw several insights:
\begin{figure}[t]
    \centering
    \includegraphics[width=0.98
\columnwidth]{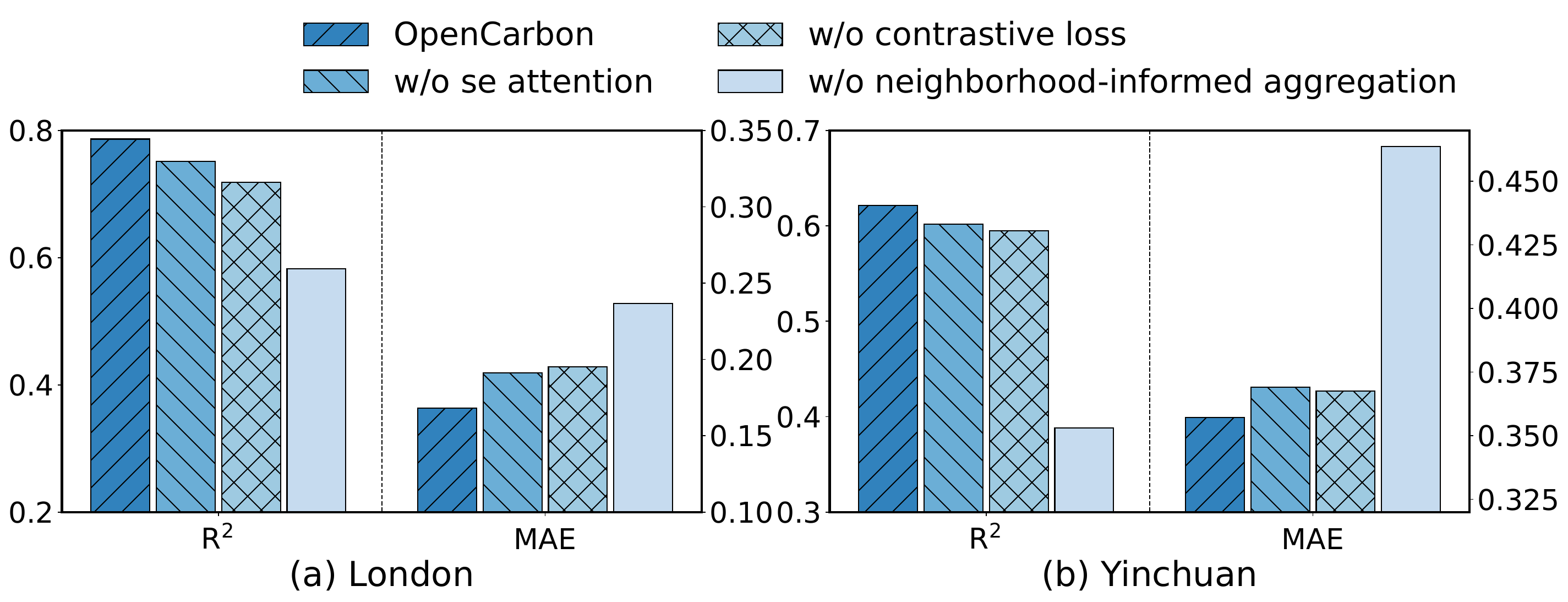}
    \vspace{-3mm}
    \caption{Ablation results on (a) London and (b) Yinchuan dataset.}
    \vspace{-4mm}
    \label{fig:ablation}
\end{figure}
\begin{itemize}[leftmargin=*]
     \item \textbf{Our model's superior performance across cities with varying levels of development.} As shown in the table, our model attains the highest performance across all three datasets, surpassing the best baseline by 28.4\%, 11.3\%, and 40.1\% in \(R^2\) for Great London, Beijing, and Yinchuan, respectively. Across the three datasets, OpenCarbon attains an average \(R^2\) of 0.6997—a level that is highly satisfactory for practical applications. These improvements demonstrate the consistent superiority of our model, regardless of the region or the data source.

     \item \textbf{Insufficient ability of existing methods in utilizing multi-modality open data to capture carbon emission's unique characteristics.} Existing carbon emission prediction methods can hardly process the satellite image modality, thereby missing out on its rich informational content. Meanwhile, current satellite image-based approaches do not effectively integrate POI distribution data with images, overlooking the interactions and complementary insights between these two modalities for modeling a grid's functionality. Therefore, existing methods fail to extract the complementary information from these two modalities to capture the functional effect of urban carbon emissions. Additionally, regarding the agglomeration effect, the poor performance of CarbonGCN demonstrates the inability of graph structures to capture the continuity of functional layouts, a task for which our neighborhood-informed agglomeration modeling module has been shown to be well-suited.

\end{itemize}

\subsection{Ablation Study}
\begin{figure}[t]
    \centering
    \vspace{-3mm}
    \includegraphics[width=0.7\columnwidth]{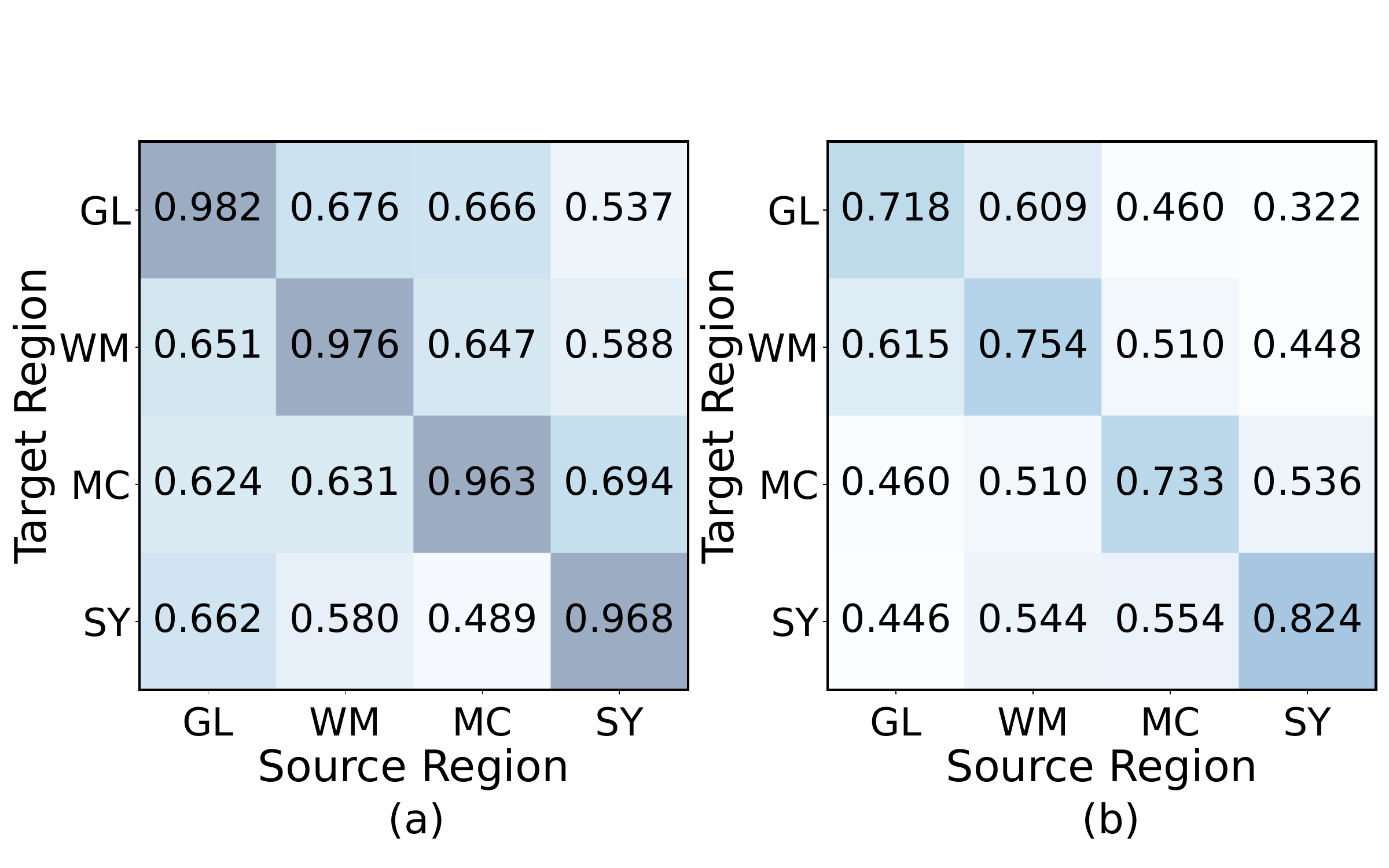}
    \vspace{-3mm}
    \caption{The Spearman’s rank correlation
coefficients for generalizability test on (a) OpenCarbon and (b) PG-SimCLR on Great London (GL), West Midlands (WM), Manchester Cities (MC), and South Yorkshire (SY).}
    \label{fig:transfer}
\end{figure}

To precisely the contribution of each design in OpenCarbon, we conduct an ablation study on one developed region, London, and one developing region, Yinchuan. Specifically, we remove the see attention and contrastive loss design in the cross-modality information extraction module, together with the neighborhood-informed aggregation module respectively. As shown in Fig~\ref{fig:ablation}, removing the SE attention module brings about a performance decrease of 3.8\% on $R^2$, indicating the necessity to consider the distinct importance of different categories of facilities. Also, removing the contrastive loss design decreases $R^2$ by 6.49\% and increases MAE by 9.49\% on average. Such a performance decrease proves the essential contribution of the contrastive loss mechanism in extracting the complementary functional semantics from both the satellite image view and the facility distribution view. Further ablation studies on the neighborhood-informed aggregation module, which brings a 31.8\% decrease of $R^2$, reveal the significant contribution of the module to characterize the agglomeration effect of emission distributions. 

\subsection{Generalizability Study}
Unlike common socioeconomic indicators, emission factors vary across regions due to differences in energy structure and accessibility, making the same activity in different areas cause different carbon emissions. Therefore, there is an inherent distribution shift in carbon emissions that hinders the generalizability of prediction techniques. However, our experimental analysis shows that despite this distributional deviation, OpenCarbon's prediction results maintain their internal rank order, allowing governors to identify high-emission areas with significant mitigation potential. To evaluate our model's ranking ability, we perform direct transfer experiments: transfer the trained model to the new region directly without further fine-tuning, and use Spearman's Rank Correlation Coefficient to assess the consistency of emission magnitudes across grids. Since POI data from our three main datasets are not category-aligned due to differences in their collection sources, we did not perform generalizability across our main datasets but instead collected data on four ceremonial counties in England for testing. As shown in Figure~\ref{fig:ablation}, OpenCarbon performed well across all twelve source-target pairs, achieving an average Spearman's rank correlation coefficient of 0.6204, outperforming PG-SimCLR by 26.56\%. This demonstrates OpenCarbon's strong generalizability in ranking emissions, highlighting its effectiveness in modeling the relationship between land use, activities, and carbon emissions.

\subsection{Case Study}
\begin{figure}[t]
    \centering
    \includegraphics[width=0.75\columnwidth]{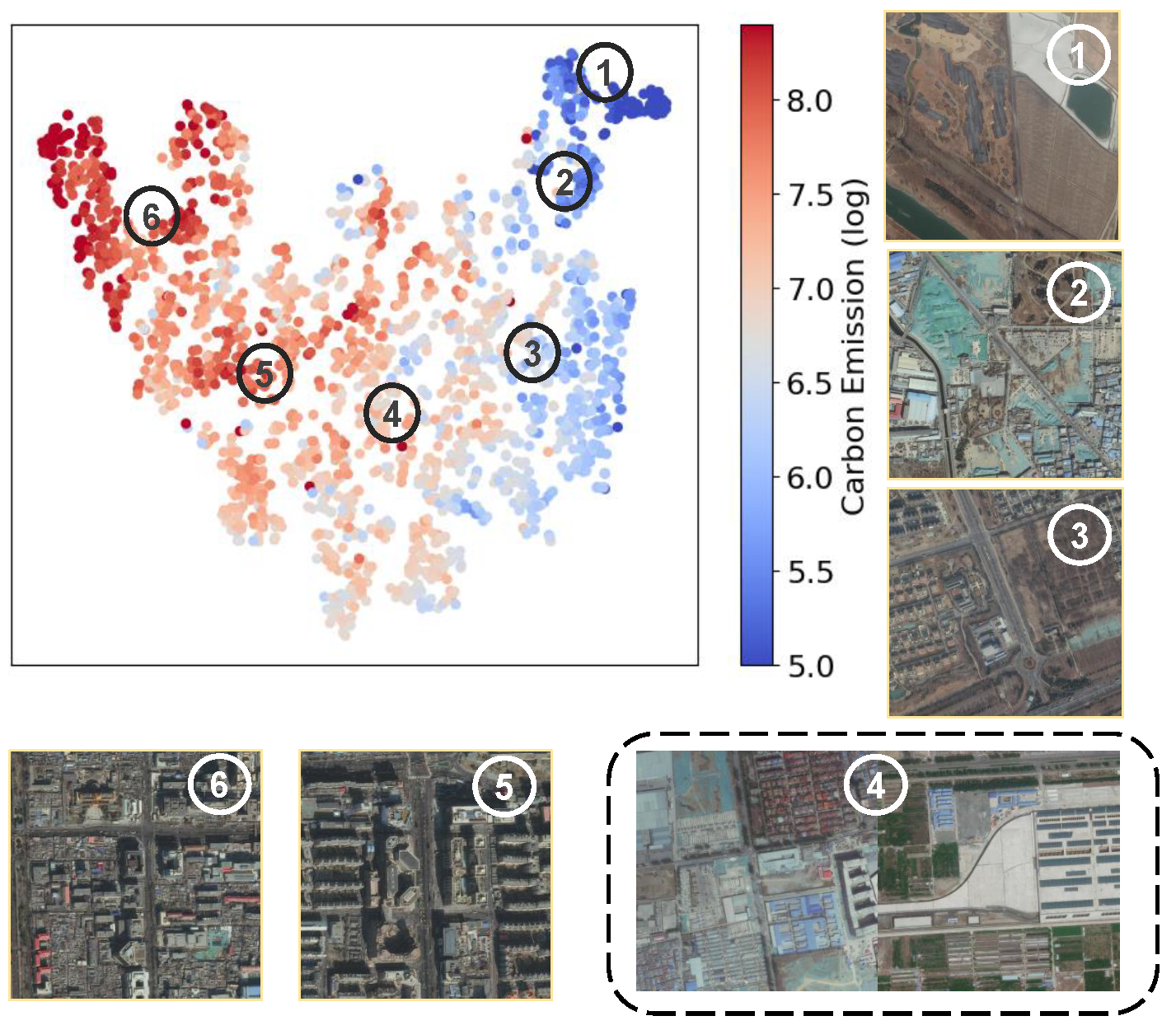}
    \caption{Visualization of the representation space. The color corresponds to the (logarithmic) grid carbon emission.}
    \label{fig:visualization}
\end{figure}

\subsubsection{Grid Representation Visualization v.s. Carbon Emission}
This section examines the distribution of grids with varying carbon emissions in the representation space. We use t-SNE~\cite{van2008visualizing} to project the representations from Equation~\ref{equ:fusion} into 2D, with dot colors indicating carbon emission levels. The direction in the representation space corresponds to increasing emissions. We select six anchor points along this direction and show a representative satellite image for each in Figure~\ref{fig:visualization}. Carbon emission levels generally correlate with land use type and density. Grids near anchor point 1 represent underdeveloped farmland and forests, while those near anchor point 2 correspond to rural areas with basic infrastructure. Anchor point 3 shows denser structures and roads, anchor point 4 includes residential and industrial zones, anchor point 5 covers high-density urban areas, and anchor point 6 highlights highly urbanized, non-residential areas. These observations highlight the link between carbon emissions and grid functionality, emphasizing the importance of incorporating functional effects into our design.

\begin{figure}[t]
    \centering
    \includegraphics[width=0.75\columnwidth]{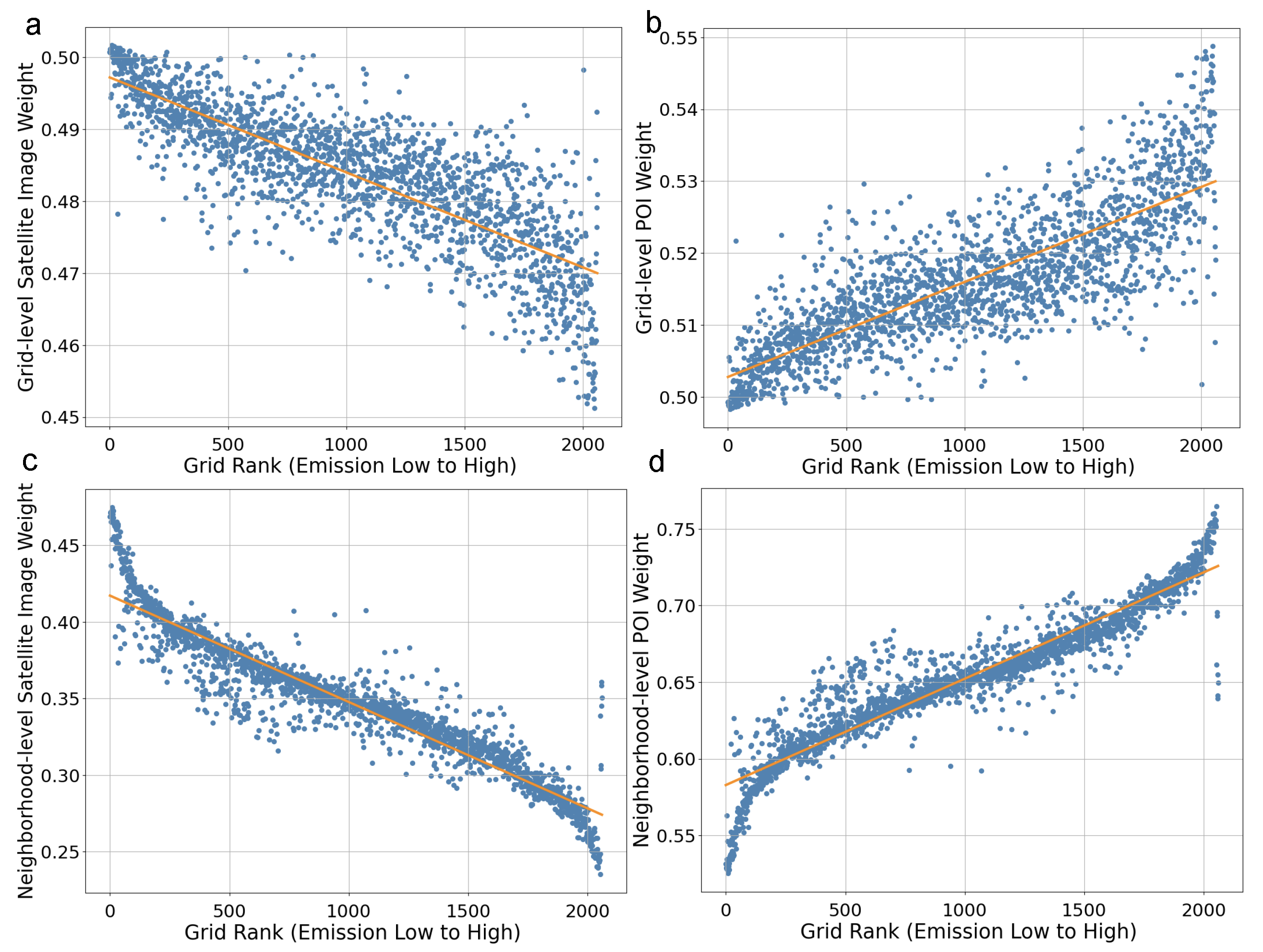}
    \caption{Visualizations of the aggregate-attention weights of the two modalities at the grid and neighborhood level. }
    \label{fig:attentions}
\end{figure}
\subsubsection{Explanatory Insights}
In OpenCarbon, we incorporate aggregate-attention to aggregate the representations of the satellite image and POI modalities. Since the aggregate-attention mechanism assigns softmaxed weights to these two representations and performs a weighted sum, the change in weight during emission rise reflects the relative importance of each modality. As shown in Figure~\ref{fig:attentions}, for grids with higher emissions, the weights for POI modalities consistently increase at both the grid level and the neighborhood level. This interesting phenomenon suggests that fine-grained functional information becomes more important for emission prediction modeling of high-emission areas, which tend to have a higher concentration of functional activities. Furthermore, one insight from this is that, while the overall layout of the city remains largely static and difficult to change, carbon emission mitigation efforts could focus on planning specific functional activities within these layouts, steering them toward more carbon-friendly options.

\section{Conclusion and Future Work}
In this paper, we introduce OpenCarbon, a neighborhood-informed attentive neural network with contrastive learning for cross-modality fusion, designed to predict high-resolution carbon emissions using open data from satellite images and POI distribution. By capturing the unique functional and spatial agglomeration effects of high-resolution carbon emissions, our model significantly outperforms existing methods for both carbon emission prediction and open data-based socioeconomic forecasting. Once trained, OpenCarbon can efficiently predict carbon emissions for new locations based on the corresponding open data inputs, greatly reducing the data collection burden of traditional carbon accounting. As a result, OpenCarbon enables targeted carbon governance and mitigation planning, driving progress toward urban sustainability. Future objectives include higher temporal resolution and prediction under resolution scaling.

\section*{Acknowledgements}
This work was supported by the National Key Research and Development Program of China under 2024YFC3307603. This work is also supported in part by Tsinghua University-Toyota Research Center. We also want to express gratitude to J. Zhao and S. Li for providing some processed data.

\bibliographystyle{named}
\bibliography{ijcai25}



\end{document}